\documentclass{article}

\usepackage[preprint,nonatbib]{neurips_data_2024}

\usepackage[utf8]{inputenc} 
\usepackage[T1]{fontenc}    
\usepackage[pagebackref,breaklinks,colorlinks,citecolor=blue]{hyperref}
\usepackage{url}            
\usepackage{booktabs}       
\usepackage{amsfonts}       
\usepackage{nicefrac}       
\usepackage{microtype}      
\usepackage{xcolor}         
\usepackage{graphicx}
\usepackage{subcaption}
\usepackage{amsmath}

\makeatletter
\newcommand{\myfnsymbol}[1]{%
  \expandafter\@myfnsymbol\csname c@#1\endcsname
}
\newcommand{\@myfnsymbol}[1]{%
  \ifcase #1
  \or \TextOrMath{\textasteriskcentered}{*}
  \or 1
  \or 2
  \or 3
  \fi
}
\newcommand{\equalcontributor}{\@myfnsymbol{1}}
\newcommand{\affiliationMBRDI}{\@myfnsymbol{2}}
\newcommand{\affiliationIIITH}{\@myfnsymbol{3}}
\newcommand{\affiliationMPII}{\@myfnsymbol{4}}
\makeatother

\title{SynthForge: Synthesizing High-Quality Face Dataset with Controllable 3D Generative Models}

\author{%
  Abhay Rawat \textsuperscript{\affiliationMBRDI,\equalcontributor} \\
  \And
  Shubham Dokania \textsuperscript{\affiliationMBRDI,\equalcontributor} \\
  \And
  Astitva Srivastava \textsuperscript{\affiliationIIITH} \\
  \And
  Shuaib Ahmed \textsuperscript{\affiliationMBRDI} \\
  \And
  Haiwen Feng \textsuperscript{\affiliationMPII} \\
  \And
  Rahul Tallamraju \textsuperscript{\affiliationMBRDI} \\
}

\begin{document}

\renewcommand{\thefootnote}{\myfnsymbol{footnote}}
\maketitle
\footnotetext[1]{Equal Contribution}%
\footnotetext[2]{Mercedes-Benz Research \& Development India \textit{(firstname.lastname@mercedes-benz.com)}}%
\footnotetext[3]{IIIT Hyderabad \textit{(astitva.srivastava@research.iiit.ac.in)}}%
\footnotetext[4]{Max-Planck Institute for Intelligent Systems, Tuebingen (\textit{haiwen.feng@tuebingen.mpg.de})}%

\setcounter{footnote}{0}
\renewcommand{\thefootnote}{\fnsymbol{footnote}}

\begin{figure}[h!]
    \centering
    \includegraphics[width=0.95\linewidth]{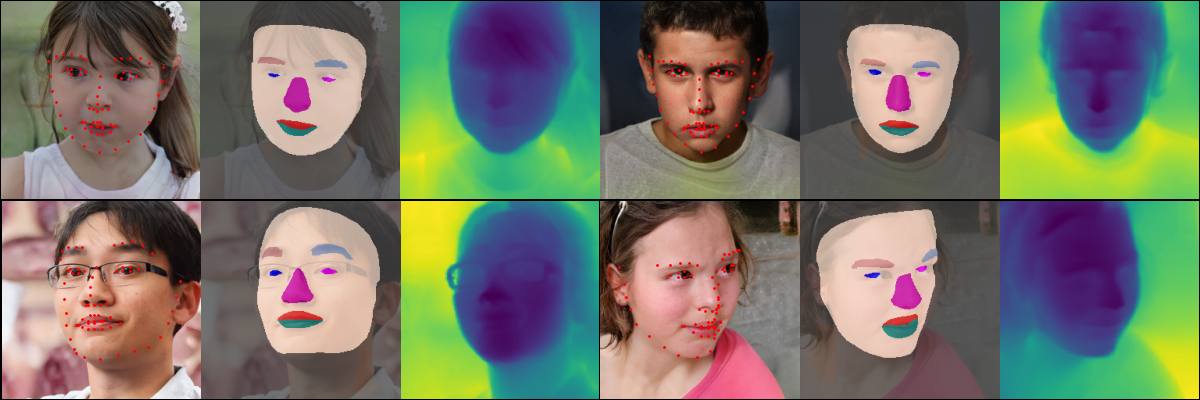}
    \caption{Our 3D-aware synthetic data pipeline produces high-fidelity images with dense multi-task spatial annotations. Compared to physically based rendering methods, we provide improvements in terms of compute, cost, and time efficiency. The generated dataset is further used for training facial analysis models for the tasks of semantic segmentation, depth estimation, and keypoint estimation.}
    \label{fig:banner}
\end{figure}%

\begin{abstract}

Recent advancements in generative models have unlocked the capabilities to render photo-realistic data in a controllable fashion.
Trained on the real data, these generative models are capable of producing realistic samples with minimal to no domain gap, as compared to the traditional graphics rendering.
However, using the data generated using such models for training downstream tasks remains under-explored, mainly due to the lack of 3D consistent annotations.
Moreover, controllable generative models are learned from massive data and their latent space is often too vast to obtain meaningful sample distributions for downstream task with limited generation.
To overcome these challenges, we extract 3D consistent annotations from an existing controllable generative model, making the data useful for downstream tasks.
Our experiments show competitive performance against state-of-the-art models using only generated synthetic data, demonstrating potential for solving downstream tasks. Project page: \href{https://synth-forge.github.io}{https://synth-forge.github.io}
\end{abstract}

\vspace{-6mm}
\section{Introduction}
\label{sec:intro}

The field of facial analysis encompasses a range of critical tasks, including recognition, expression analysis, and biometric authentication, each dependent on accurate face parsing tasks.
As the demand for robust facial analysis systems escalates, the need for extensive and diverse datasets to train these systems becomes paramount.
Traditional data collection and labeling methods are not only time-consuming and costly but also susceptible to human bias and errors.
An exponential rise in data demand, fueled by the digital transformation across industries, has incited the need for innovative and efficient data acquisition techniques.
Recent advancements in synthesizing training data for face analysis have garnered considerable attention, pointing to a significant shift in the methodologies employed for training facial recognition and analysis models.
Unlike traditional data, which relies on real-world occurrences for collection, synthetic data is algorithmically generated, offering a plethora of advantages in terms of cost, scalability, and accuracy.

Traditional approaches, heavily reliant on sophisticated computer graphics pipelines \cite{fitymi,hodan2019photorealistic,hodavn2020bop}, offer the ability to render high-quality, auto-annotated facial images.
However, the complexity and prohibitive cost of setting up such pipelines limit their accessibility and scalability \cite{fitymi,hodavn2020bop,hodan2019photorealistic}. 
Additionally, a persistent challenge with models trained on physically based rendering (PBR) synthesized imagery is the domain gap — the discrepancy between synthetic and real-world data distributions — which necessitates elaborate domain adaptation techniques to achieve practical utility.


In contrast, the emergence of advanced generative models promises a new horizon in the generation of photo-realistic human faces.
However, their potential for downstream face analysis tasks is yet to be fully harnessed, primarily due to the difficulty in obtaining 3D consistent annotations for the generated images \cite{dadnet++}. 
Synthetic data obtained from generative models is particularly well-suited for providing such comprehensive supervision for computer vision tasks as has been seen in the community \cite{fitymi,Kar_2019_ICCV,varol17_surreal,gomez2019large}, however yet remains under-explored in the facial analysis domain with 3D consistent annotations for the corresponding generated data.
Controllable generative models are usually trained on real data, and are therefore capable of generating realistic data with minimal or no domain gap.
However, the space spanned by the latent control variables of these models are generally too vast to obtain meaningful samples.
Uniformly sampling the latent space for data generation would result in samples that might not represent the distribution required to meaningfully train for a downstream task, and often provide redundant information that contributes towards scaling the dataset size without significant enhancement to task performance.

The objective of this research is to mitigate the shortcomings related to the use of controllable generative models for synthetic dataset curation, towards training downstream tasks more efficiently.
To address the inherent challenges, we introduce two key insights: (i) the utilization of a generative model for the synthesis of high-quality facial images, alongside 3DMMs for extracting 3D-aware annotations, and (ii) the employment of this exclusively synthesized and annotated dataset for training a robust multi-task model capable of predicting keypoints (kp), semantic segmentation (seg), and depth modalities
As the annotation schema obtained from the generative model differ from those followed in the existing benchmark datasets, we adopt a label fine-tuning strategy.
Label finetuning is aimed at enhancing the predictions of our synthetically trained network to match those of the real-world datasets \cite{CelebAMask-HQ,liu2020new,karras2019style,sagonas2013300,richter2016playing}.
Although the label fine-tuning strategy enhances the model's performance for evaluation purposes on existing datasets, it is primarily an evaluative/regulative step.
The downstream models, as trained exclusively with synthetic data, demonstrate strong performance on real-world data directly, underscoring its practical viability and the effectiveness of our pipeline.

\begin{figure*}[t]
  \centering
  \includegraphics[width=0.98\linewidth]{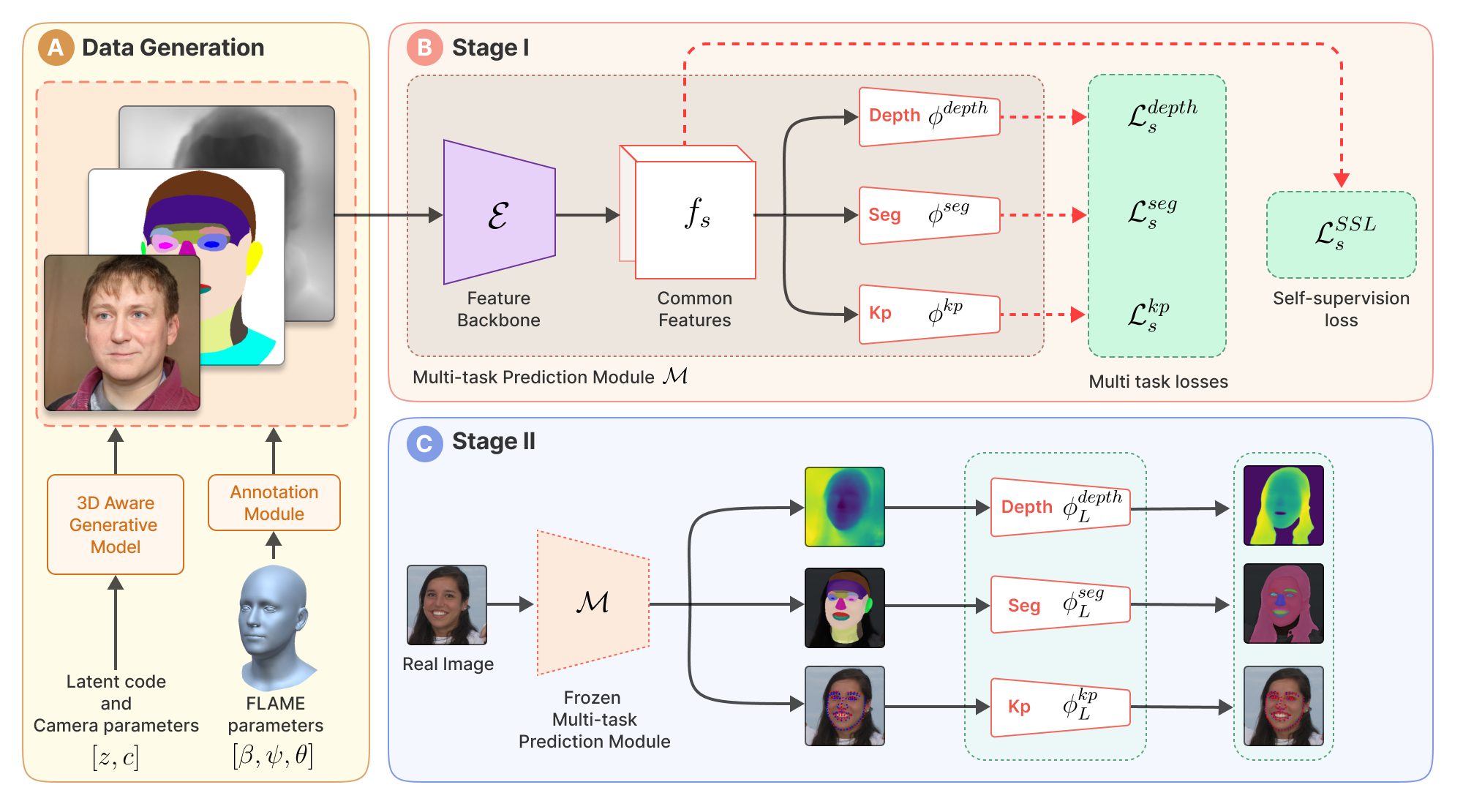}
  \caption{This figure illustrates a comprehensive framework for facial analysis leveraging synthetic data. (A) involves generating a synthetic dataset using a 3D-aware generative model, with annotations derived from FLAME parameters, to (B) train a multi-task prediction module ($\mathcal{M}$) with self-supervision loss in Stage I. (C) Stage II utilizes the pre-trained module to fine-tune predictions on real-world data.}
  \label{fig:overall_process}
\end{figure*}

A schematic of our method for obtaining multi-modal annotations is depicted in figure \ref{fig:overall_process}, along with image samples of the generated data in figure \ref{fig:banner} showcasing the practical outcomes of our proposed pipeline. The key contributions in the proposed work are as follows:

\begin{itemize}
    \item \textbf{Annotated Data Generation in a 3D-aware pipeline}: Utilizes a 3DMM-controllable generative model to extract soft semantic and spatial labels across a comprehensive range of facial features, providing dense, multi-modal supervision essential for detailed facial analysis and parsing.
    \item \textbf{Multi-Task Framework for Synthetic Data}: Employs a synthesized dataset to train a robust multi-task model that accurately predicts various facial attributes such as keypoints, segmentation, and depth, showcasing the potency of synthetic data in enhancing complex model training.
\end{itemize}

Generative models show potential in overcoming the domain gap between synthetic and real-world data while providing a scalable and cost-effective alternative to traditional data collection and annotation methods for facial analysis.
The cost of acquiring the data from these generative models can be reduced further by efficiently sampling the latent space for the data generation process guided by the down-stream tasks.
In conclusion, we show that models trained on the controllable generative synthetic data are on par with the existing state-of-the-art on various facial analysis benchmarks. The project page is available at \href{https://synth-forge.github.io}{https://synth-forge.github.io} highlighting the links to additional resources, released dataset and code for reproducability.

\section{Related Works}
\noindent
\textbf{Datasets for Facial Analysis:} 
Deep learning methods that aim at facial analysis, such as face detection \cite{li2018dsfd, deng2019retinaface} and face parsing \cite{lin2021roi, Sarkar_2023_CVPR}, require high-quality face images with reliable \& accurate task-specific annotations, e.g. facial landmarks or semantic labels.
Several datasets \cite{le2012interactive, CelebAMask-HQ, liu2020new} have been proposed in this regard, which contain tens of thousands of face images. More specifically, CelebAMask-HQ\cite{CelebAMask-HQ} contains 30,000 high-resolution face images selected from the CelebA\cite{liu2015faceattributes} dataset by following CelebA-HQ\cite{karras2017progressive}. All the images have manually annotated segmentation masks of facial attributes of 512 x 512 resolution, spread among 19 classes including all facial components and accessories. Another dataset LaPa\cite{liu2020new} comprises over 22,000 facial images showcasing a wide array of expressions, poses, and occlusions. Each image in LaPa comes with an 11-category pixel-level label map and 106 facial landmark points. 

Though such datasets seem large-scale, they are often limited in terms of scalability and diversity, along with being resource intensive, and subject to human errors through manual annotations. To bridge this gap, \cite{fitymi} proposes a computer graphics inspired PBR-based synthetic data generation pipeline, procedurally generating 100,000 synthetic faces with precise 2D landmarks and per-pixel segmentation labels, allowing for vast exploration in training large-scale models on downstream tasks and real-world label adaptation. While the dataset in \cite{fitymi} boast high-fidelity and large scale nature, it is very expensive to build such a dataset owing to the cost of design, compute, and the domain-gap arising from the inherent nature of PBR-based methods.
%
%
%
%
%
%
%
%
%
%
\\
\noindent
\textbf{Generative Approaches for Face Synthesis:} The progress in generative adversarial 
 learning in the past decade \cite{isola2018imagetoimage, karras2019style} has led researchers to introduce learning-based methods for generating training data. \cite{jahanian2022generative} proposes a generative framework to enable data generation for multi-view representation learning and concludes that the representations learned using generated data outperform those learned directly from real data. In the context of face synthesis, methods like \cite{HUANG2023272} propose a bi-directional method for face editing by manipulating semantic maps. A recent trend is to condition the face generation by learning an inherent 3D prior in an unsupervised setup, proposed in EG3D\cite{eg3d}. This conditioning only helps in improving the generation quality \& view consistency, however, it doesn't provide any explicit control over the generation process, e.g. manipulating facial expressions. Recent advancements in diffusion models \cite{ho2020denoising, rombach2022highresolution} have given rise to text-driven generation of high-quality and highly diverse images, with methods like \cite{zhang2023adding} provide different types of control over the denoising process. \cite{ding2023diffusionrig} leverages the parametric head model to control facial image synthesis. The usage of the parametric head model enables control over facial expressions, head pose, and even lighting, however, it doesn't allow fine-grain control over hair, beard etc. Omniavatar\cite{xu2023omniavatar} address this issue by combining both parametric as well as neural implicit head representation, where the parametric head model allows controllable generation and implicit representation models of other details such as hair, beard, accessories etc. Another similar and recent work, Next3D\cite{next3d}, built on top of EG3D utilises a FLAME \cite{flame} mesh explicitly to generate a texture map and convert to Triplanar representation, followed by volumetric rendering similar to EG3D. Although these approaches generate highly photorealistic and diverse images, a caveat is the lack of control over generation of  dense annotations.  \cite{dadnet++} addresses this issue by using a PTI-inversion \cite{roich2022pivotal} over the StyleGAN2 \cite{stylegan2} latent space in the EG3D pipeline to generate multi-view data and 3D landmarks for further training landmark estimation task. However, the lack of further dense annotations still limits the utility of existing approaches. Through the proposed pipeline in our work, we aim to bridge the gap between synthetic data generation with control over annotations, and the ease of complexity of synthetic data generation. 
 \\
 \noindent
\textbf{ Multi-task Facial Analysis:}
While there exist datasets in the facial analysis domain with annotations for multiple tasks, the exploration in the domain of multi-task learning for face parsing and analysis is very recent. \cite{lee2021face} provide a framework to mutually learn semantic segmentation and depth information through domain translation, and also release the LaPa-D dataset. \cite{zheng2021farl} explores the domain of multi-task learning through dense pre-training on text-image pairs to learn robust representations, and then fine-tunes on face parsing, alignment, and attribute recognition.; However, The success of the approach can be attributed to the extensive pre-training on the text-image dataset, which may not be easily available on a large-scale. The work presented in \cite{zheng2022decoupled} performs a cyclic self-regularization for face parsing tasks by utilizing related tasks such as edge estimation and edge categorization. The caveat with such approaches is the requirement of dense multi-task datasets which are resource intensive to compile. Multi-task approaches using partially annotated data allow for handling of sparsely annotated datasets in a multi-task manner, such as the approach shown in \cite{mtpsl}. We follow a similar methodology towards multi-task learning with cross-task feature representations combined with the capability to generate large volumes of densely annotated synthetic data. The proposed approach towards multi-task learning allows for leveraging the relationships between tasks to extract richer feature representaions in the downstream tasks

\vspace{-3mm}


\section{Method}
\label{sec:method}

Our research aims to enhance multi-task facial analysis, with a focus on synthetic data. We introduce a three-component generative pipeline: (A) Synthetic data generation employing a 3D-aware generative model with an annotation module, (B) Training a multi-task network on synthetic data, and (C) Real-world label fine-tuning for evaluation on standard real-world benchmarks. In \textbf{Data Generation (A)}, detailed in \ref{compA}, we use Next3D \cite{next3d} and FLAME \cite{flame} to generate images and annotations for facial attributes. \textbf{Synthetic Backbone (B)} involves training a network with self-supervised learning, which is later frozen for Stage-II training (see \ref{compB}). \textbf{Label Finetuning (C)}, as elaborated in \ref{compC}, fine-tunes this network prediction head on real images, fine-tuning each task's predictions for real-world application. This approach seeks to boost the accuracy and adaptability of facial analysis tasks.

\subsection{Data Generation Pipeline}
\label{compA}

We incorporate Next3D \cite{next3d} as the 3D-aware generative model which uses FLAME as the parametric head model (3DMM), allowing a disentangled control of the facial geometry and expression in the generated image. The generative approach in Next3D models dynamic and static components with two independent tri-plane branches. The first branch is a generative texture-rasterized tri-plane which operates on a view-dependent synthesized texture map from a StyleGAN2 \cite{stylegan2} network. The inputs are a set of randomly sampled latent vector $z \sim \mathcal{N}(0, I)$ and the camera parameters for the synthesized view $c \in \mathbb{R}^{1\times25}$, where $c = K \cdot [R | t]$ represents the camera intrinsic and extrinsic parameters. The generated textures are applied to a deformed FLAME head mesh which is parameterized by the shape, expression, and pose parameters denoted as $\beta$, $\psi$, and $\theta$ respectively. The second branch models the static components into a tri-plane feature map following the method proposed in EG3D \cite{eg3d}. The tri-plane features from both the branches are fused through a neural blending module and used for neural volume rendering, followed by an image super-resolution module to generate realistic face images, $\mathcal{I}^{synth} \in \mathbb{R}^{512 \times 512 \times 3}$.

\begin{figure}[h]
    \centering
    \begin{subfigure}{0.25\columnwidth}
        \includegraphics[width=\columnwidth]{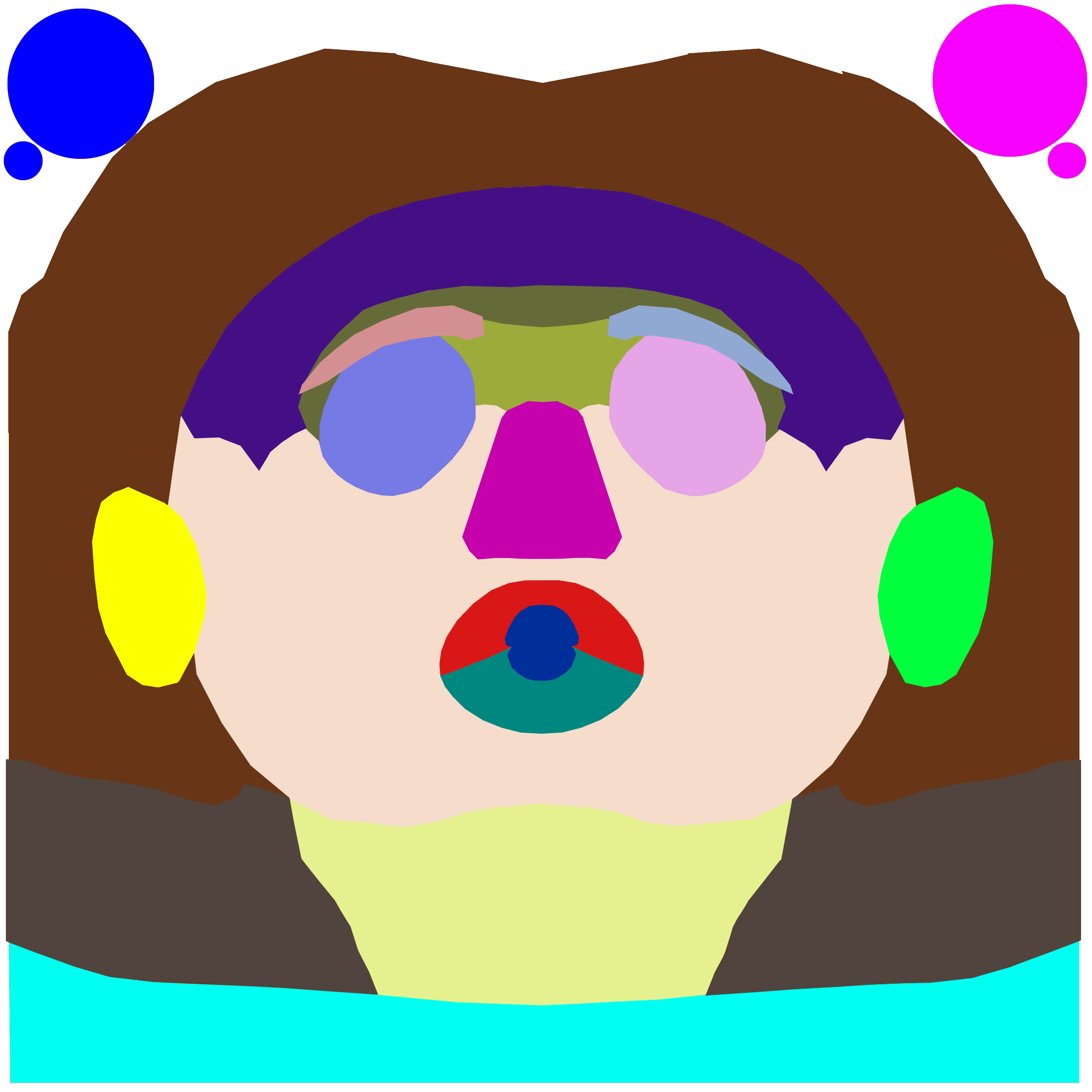}
        \caption{Flame texture}
        \label{fig:texture}
    \end{subfigure}
    \begin{subfigure}{0.25\columnwidth}
        \includegraphics[width=\columnwidth]{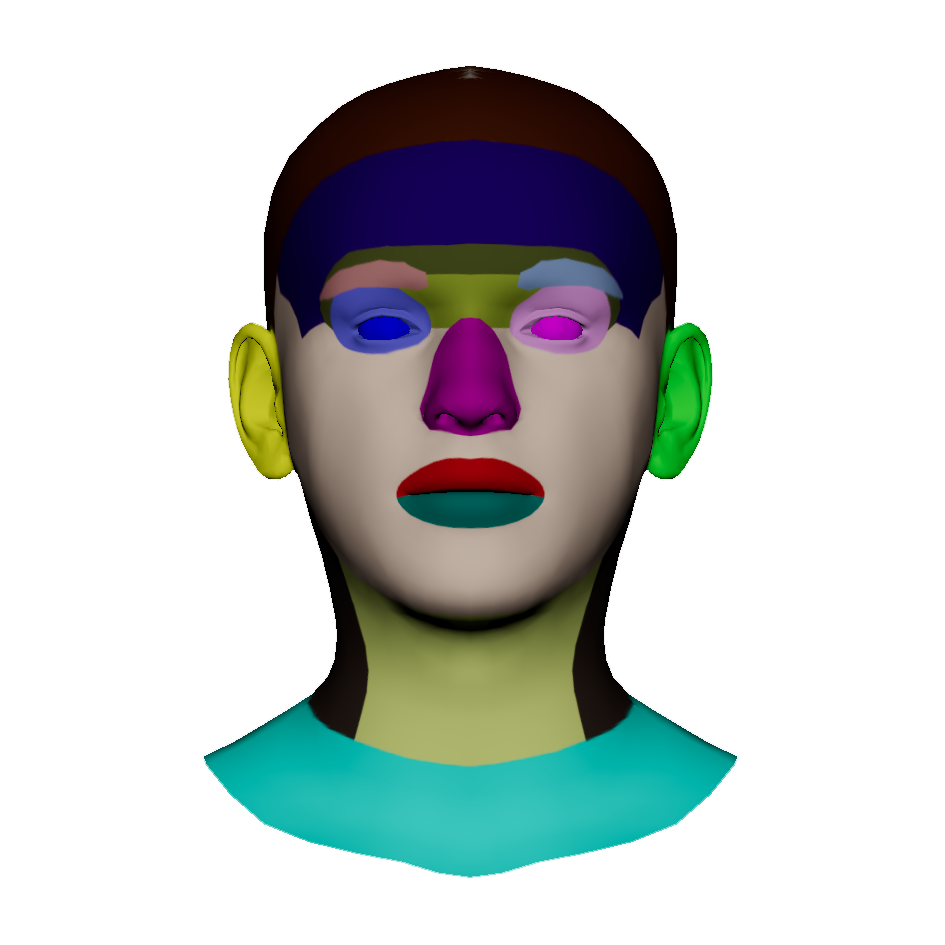}
        \caption{Flame texture on head}
        \label{fig:texture_head}
    \end{subfigure}
    \caption{UV Texture map for the FLAME head model showing semantic labels which are used to generate the segmentation annotations for our proposed data generation pipeline. These segmentation masks can be updated through the UV map to adapt to finer or coarser level of semantic information.}
    \label{fig:flame_seg}
\end{figure}

In our pipeline, annotations derive from Next3D's neural rendering and FLAME mesh transformations. Semantic segmentation and 68 3D landmarks from FLAME mesh's ground truth are projected onto generated images, utilizing camera parameters to align with Next3D's transformations, resulting in accurate 3D and 2D landmark representations as depicted in fig. \ref{fig:flame_seg} and fig. \ref{fig:overall_process} (A). Minor misalignments exist in the transformation process due to the non-linear transformations through the StyleGAN modules in the Next3D pipeline. Towards the rectification and alignment of the generated annotations, we utilise an \textit{alignment process} for the FLAME mesh which optimizes the consistency between the 3D mesh points and neural densities from the Next3D rendering module. More details and exploration about the \textit{alignment process} are discussed in the Supplementary material. Depth maps are consistently aligned with RGB images already, owing to volumetric rendering based on NeRF\cite{nerf}. This approach enables the creation of a diverse, large-scale annotated dataset, enhancing the training efficiency for subsequent stages.
    
\subsection{Multi-Task Synthetic Backbone (Stage I)}
\label{compB}

In \textit{Stage I}, we train a multi-task model on the synthetically generated data. This process is guided by a training regimen that capitalizes on the dense multi-task annotations provided by our data generation pipeline.
\vspace{-3mm}
\paragraph{Model Formulation:} Following common multi-task learning paradigms \cite{vandenhende2021multi,liu2019end}, the pipeline starts with the input image $x_s \in \mathbb{R}^{3 \times H \times W}$, representing the synthetically generated image $\mathcal{I}^{synth}$ in this case. This image undergoes initial processing through a feature encoder block $\mathcal{E}: \mathbb{R}^{3 \times H \times W} \to \mathbb{R}^{C \times H \times W}$, to generate the feature map $f_s = \mathcal{E}(x_s) \in \mathbb{R}^{C \times H \times W}$, where $C$ represents the number of feature channels, is the basis for further task-specific processing. Each task's processing is handled by individual task heads $\phi^{t_i}:\mathbb{R}^{C \times H \times W} \to \mathbb{R}^{C' \times H' \times W'}$. For our tasks, these heads are $\phi^{seg}$, $\phi^{depth}$, and $\phi^{kp}$, with their corresponding predictions denoted as $\hat{y}^{t_i}_s = \phi^{t_i}(f_s)$.
\vspace{-3mm}
\paragraph{Training Losses:} The training for each task utilizes specific loss functions: L1-norm loss for depth estimation ($\mathcal{L}^{depth}_s$), cross-entropy loss for semantic segmentation ($\mathcal{L}^{seg}_s$), and L2-norm loss for keypoint estimation ($\mathcal{L}^{kp}_s$). The overall task loss is defined as:
\vspace{-2mm}
\begin{equation}
    \mathcal{L}_s^{task} = \frac{1}{N} \sum\limits_{n=0}^N \sum\limits_{t_i \in \mathcal{T}} \lambda_{s,n}^{t_i} \mathcal{L}_s^{t_i}(\hat{y}^{t_i}_{s,n}, y^{t_i}_{s,n})
    \label{eq:task_loss}
\end{equation}
Here, $N$ is the number of samples, $\mathcal{T} \in \{seg, depth, kp\}$ represents the set of tasks, $\lambda_{s,n}^{t_i}$ is the loss weight for task $t_i$, and $y^{t_i}_{s,n}$ is the ground truth label for the task. Additionally, to enhance feature representation in the backbone network, we incorporate a self-supervised loss term,  based on an affine transformation $\epsilon$ applicable to both the input and the intermediate feature. The SSL loss term is formulated as:
\vspace{-2mm}
\begin{equation}
    \mathcal{L}^{SSL}_s = \frac{1}{N} \sum\limits_{n=0}^N ||\mathcal{E}(\epsilon(x_{s,n})) - \epsilon(\mathcal{E}(x_{s,n}))||_{_2}
\end{equation}
Upon completion of the training, we freeze the weights of the multi-task synthetic backbone network for use in the next stage. The network in its frozen state is denoted as $\mathcal{M}$, and its predictions are $\hat{y}^{seg}_s$, $\hat{y}^{depth}_s$, and $\hat{y}^{kp}_s$ for the respective tasks. 


\subsection{Multi-Task Label Fine-tuning (Stage II)}
\label{compC}

\textit{Stage II} of our methodology focuses on label fine-tuning, a crucial process for adapting the multi-task network's predictions to various real-world datasets for fair evaluations. This stage utilizes the network trained in Stage I and fine-tunes its predictions to align with real-dataset annotations. The input to the pre-trained multi-task synthetic backbone network, denoted as $\mathcal{M}$, is $x_r \in \mathbb{R}^{3 \times H \times W} \in \mathcal{I}^{real}$, from the real-world dataset. Processing this input through $\mathcal{M}$, we obtain predictions for segmentation, depth, keypoints, and the intermediate feature map,  $(\hat{y}^{seg}_s, \hat{y}^{depth}_s, \hat{y}^{kp}_s, f_s) = \mathcal{M}(x_r)$.

\paragraph{Task-Specific Label Fine-Tuning Networks:} For each task $t_i \in \mathcal{T}$, we construct specific label fine-tuning networks, $\phi^{t_i}_L$, such as $\phi^{seg}_L$, $\phi^{depth}_L$, and $\phi^{kp}_L$, based on the tasks available in the real-world dataset. The predictions from these networks are formulated as $\hat{y}_r^{t_i} = \phi^{t_i}_L(\hat{y}_s^{t_i}, f_s)$, for each task $t_i \in \mathcal{T}$. The inclusion of the intermediate feature map $f_s$ in the fine-tuning process is optional and subject to exploration in our ablation studies. This approach is designed to account for the label disparity between synthetic and real annotations, accommodating attributes like hair and accessories that may not be present in the generated data but are found in real-world datasets.

The training of stage II follows a similar routine as in Stage I for loss metrics. The implementation of this label fine-tuning stage significantly enhances the model's versatility and utility across diverse datasets. This approach not only aligns the model's predictions with real-world data but also ensures that the model retains its robustness and accuracy, thereby elevating its overall performance in practical applications.

\begin{table}[t]
\centering
\begin{tabular}{@{}lcccr@{}}
\toprule
Method                                      & Common   & Challenge  \\ \midrule
Wood et al. \cite{fitymi}                         & 5.61     & 8.43       \\
Wood et al. \cite{fitymi} + Label Adaptation      & 3.09     & 4.86       \\
FaRL-B \cite{zheng2021farl}                              & \textbf{2.50}     & 4.42       \\
STAR  \cite{star}                               & 2.52     & \textbf{4.32}       \\
DECA \cite{deca} + LF                   & 4.26     & 7.33       \\ 
DECA \cite{deca} + LF w/ $f_s$          & 3.86     & 6.40       \\
\midrule
SB                                      & 5.74     & 9.30       \\
Random Sampling $10K$ + LF              & 5.18     & 9.25       \\
Random Sampling $10K$ + LF w/ $f_s$     & 5.06     & 8.84       \\
SB + LF                                 & 3.01     & 4.96       \\ 
SB + LF w/ $f_s$                        & \textbf{2.90}     & \textbf{4.74}       \\  \bottomrule
\end{tabular}
\caption{Table for results on Keypoints to compare single task performance of proposed method with current SOTA methods. SB and LF refere to Synthetic Backbone and Label finetuning respectively. $f_s$ are the features extracted from the feature encoder $\mathcal{E}$ trained on the synthetic data.}
\label{tab:300w}
\end{table}




\begin{table}[t!]
\centering
\resizebox{\textwidth}{!}{%
\begin{tabular}{@{}lccccccccccc@{}}
\toprule
\textbf{Method}                 & \textbf{Skin}  & \textbf{Hair}  & \textbf{L-eye} & \textbf{R-eye} & \textbf{U-lip} & \textbf{I-mouth} & \textbf{L-lip} & \textbf{Nose}  & \textbf{L-brow} & \textbf{R-brow} & \textbf{Mean}  \\ \midrule
Liu et al.  \cite{liu2020new}               & 92.7  & 96.3  & 88.1  & 88.0  & 84.4  & 87.6    & 85.7  & 95.5  & 87.7   & 87.6   & 89.8  \\
Te et al. \cite{te2020edge}                 & 97.3  & 96.2  & 89.5  & 90.0  & 88.1  & 90.0    & 89.0  & 97.1  & 86.5   & 87.0   & 91.1  \\
Wood et al. \cite{fitymi}                   & 97.1  & 85.7  & 90.6  & 90.1  & 85.9  & 88.8    & 88.4  & 96.7  & 88.6   & 88.5   & \textbf{90.1}  \\
FaRL-B  \cite{zheng2021farl}                & 98.00 & 96.52 & 93.97 & 93.91 & 90.15 & 91.74 & 91.21 & 97.92 & 92.70 & 92.65 & \textbf{93.88} \\ \midrule
RS + LF $10k$ samples                                    & 96.53 & 94.26 & 86.12 & 85.58 & 81.36 & 85.48   & 85.74 & 95.68 & 85.54  & 84.48  & 87.98 \\
DECA + LF $100k$ samples                                 & 96.36 & 93.69 & 86.20 & 86.05 & 81.89 & 84.64   & 82.74 & 95.50 & 86.61  & 85.74  & 87.94 \\
SB + LF $100k$ samples                                      & 96.80 & 94.30 & 88.60 & 88.60 & 84.43 & 87.97   & 87.13 & 96.16 & 87.64  & 88.62  & \textbf{90.03} \\ \bottomrule
\end{tabular}}
\caption{Comparison with state of the art face parsing methods on the LaPa. RS, SB and LF refer to Random Sampling, Synthetic Backbone and Label Finetuning respectively. $f_s$ are the features extracted from the feature encoder $\mathcal{E}$ trained on the synthetic data.}
\label{tab:lapa}
\end{table}


\begin{table*}[]
\centering
\resizebox{\textwidth}{!}{%
\begin{tabular}{@{}lccccccccccccccr@{}}
\toprule
\textbf{Method}  & \textbf{Face} & \textbf{Nose} & \textbf{Glasses} & \textbf{LE} & \textbf{RE} & \textbf{LB} & \textbf{RB} & \textbf{L-ER} & \textbf{R-ER} & \textbf{IM} & \textbf{UL} & \textbf{LL} & \textbf{Hair} & \textbf{Neck} & \textbf{Mean} \\ \midrule
Wei et al. \cite{wei2019accurate}      & 96.4          & 91.9          & 89.5             & 87.1        & 85          & 80.8        & 82.5        & 84.1          & 83.3          & 90.6        & 87.9        & 91.0          & 91.1          & 88.1          & 87.80      \\
EAGR    \cite{te2020edge}         & 96.2          & 94.0            & 92.3             & 88.6        & 88.7        & 85.7        & 85.2        & 88            & 85.7          & 95.0        & 88.9        & 91.2        & 94.9          & 89.4          & 90.27      \\
AGRNet   \cite{te2021adaptive}        & 96.5          & 93.9          & 91.8             & 88.7        & 89.1        & 85.5        & 85.6        & 88.1          & 88.7          & 92          & 89.1        & 91.1        & 95.2          & 89.9          & \textbf{90.37}      \\
DML-CSR  \cite{zheng2022decoupled}         & 95.7          & 93.9          & 92.6             & 89.4        & 89.6        & 85.5        & 85.7        & 88.3          & 88.2          & 91.8        & 87.4        & 91.0        & 94.5          & 89.6          & 90.23         \\
FaRL-B   \cite{zheng2021farl}        & 96.29         & 93.72         & 93.91            & 88.75       & 88.64       & 85.24       & 85.46       & 87.06         & 87.36         & 90.96       & 87.53       & 89.81       & 95.6          & 91.54         & 90.13         \\ \midrule
SB + LF w/ $f_s$ & 95.6          & 93.0            & 89.13            & 86.61       & 86.32       & 83.75       & 83.61       & 84.88         & 84.57         & 89.04       & 85.85       & 88.34       & 94.03         & 89.36         & \textbf{88.15}         \\ \bottomrule
\end{tabular}
}
\caption{Comparison with state of the art face parsing methods on CelebAMask-HQ dataset. SB, LF, $f_s$ are explained in Table \ref{tab:lapa}. LE and RE refer to the left and right ear. LB and RB refer to the left and right eyebrows. L-ER and R-ER refer to the left and right ear and, IM, UL and LL refer to inner mouth, upper and lower lips respectively. Note that we do not evaluate on facial accessories.}
\label{tab:celeb}
\end{table*}

\section{Experiments \& Results}
\label{sec:results}
\subsection{Experiment Setup}
\paragraph{Model Architecture:} We employ the UNet \cite{unet} architecture for the multi-task synthetic backbone model $\mathcal{E}$ and also for the individual heads for the depth and face parsing heads - $\phi^{depth}$ and $\phi^{seg}$.
For predicting the facial landmarks, we make use of the a $2$ layer convolution network followed by a stacked hourglass architecture \cite{newell2016stacked} with $2$ stacks - which we denote as $\phi^{kp}$ in the text.
For label finetuning, we follow the same architecture for the respective tasks.

\paragraph{Data:} The data used to train the multi-task prediction module $\mathcal{M}$ is generated at a resolution of $512 \times 512$.
We generate a dataset of 100,000 samples with varying identities and expressions by i.i.d. sampling the FLAME shape $\beta$ and expression parameters $\psi$.
To ensure that our synthetically generated data contains images from different viewpoints, we position the camera by sampling its azimuth and elevation angles from a range of $[-\pi/4, \pi/4]$.
The field of view for the camera is sampled uniformly within a range of 12$^{\circ}$ to 27$^{\circ}$ and its distance from the subject is kept at a constant $2.7$ meters. Unlike physics-based rendering methods, our data generation pipeline takes a fraction of the time to generate photo-realistic data with annotations to be leveraged for training.
Using our data generation pipeline, it took about $7$ hours to generate a dataset of 100$k$ images with 3 different annotation modalities on a single NVIDIA V100 GPU with a batch size of $1$.

We use Next3D as the 3D aware generative model for our work which features disentangled control for facial geometry via FLAME.
However, this generative model can be replaced by any existing 3D aware generative model that has the potential to leverage the 3D facial geometry to obtain facial (or presumably other kinds of) annotations.
FLAME does not have access to the textures therefore we cannot obtain perfect annotations for certain categories like eyebrows and lips.
Therefore, we leverage FLAME's texture maps \cite{flame} to manually annotate these categories onto the UV texture map that we use for segmentation - Fig. \ref{fig:flame_seg}.

\paragraph{Training:}
During training, we include data augmentations like affine transformations, perspective warping, random flipping, blurring, random erasing, adding noise, and, varying the brightness and contrast adjustments.
The images are resized to a resolution of $256 \times 256$ for all our experiments.
The multi-task prediction module $\mathcal{M}$ during stage I, is trained entirely on the synthetically generated data.
For stage I, we use a learning rate of $1e-3$ for all our task heads ($\phi^{kp}$, $\phi^{depth}$ and $\phi^{seg}$) and $1e-4$ for the feature encoder $\mathcal{E}$.

During label-finetuning (stage II), the entire multi-task module $\mathcal{M}$ is frozen.
We pass the predictions of the output heads $\hat{y}^{t_i}_s$, optionally combining the feature maps obtained from the feature encoder $\mathcal{E}$, to their respective label-finetuning networks $\phi^{t_{i}}_L$.
The reason we pass the feature maps to the label-finetuning networks is to make the model aware of facial attributes like hair, eyebrows, and lips.

The learning rate in all our experiments is annealed using a cosine decay schedule with a minimum learning rate of $1e-6$.
We use AdamW optimizer to train all our models and our experiments are implemented in PyTorch \cite{Falcon_PyTorch_Lightning_2019} and Pytorch 3D \cite{ravi2020pytorch3d}.
We use Pytorch Lightning \cite{DBLP:journals/corr/abs-1711-05101} to manage and scale our experiments.

\begin{figure*}[t!]
  \centering
  \includegraphics[width=0.98\linewidth]{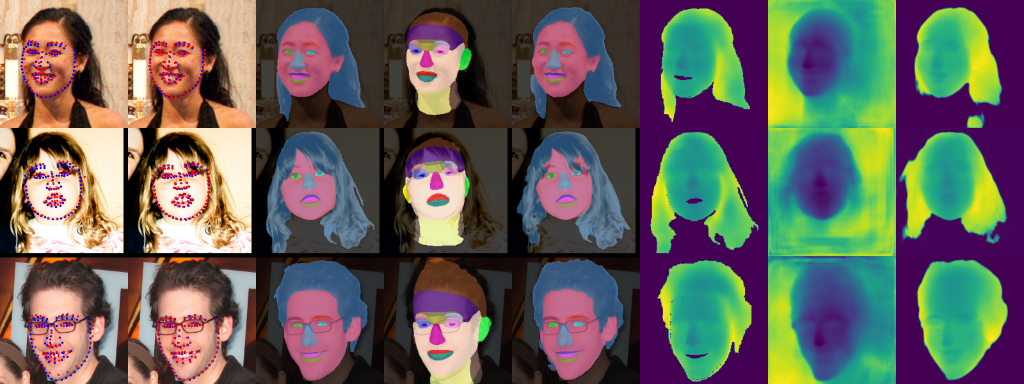}
  \caption{Qualitative results on face alignment, face parsing and depth estimation task. From left to right: a) Ground truth (GT) landmarks (blue) with predicted landmarks from SB model, b) GT landmarks (blue) with predicted SB + LF w/ $f_s$ landmarks\, c) GT face parsing, d) predicted face parsing with SB , e) predicted face parsing with SB + LF + w/ $f_s$, f) GT depth map, d) predicted depth map with SB , e) predicted depth map with SB + LF + w/ $f_s$    }
  \label{fig:lapa_res}
\end{figure*}

\subsection{Datasets}
\textbf{Facial Landmark Localization:} We evaluate our approach on the 300W dataset which is a popular landmark localization benchmark.
We note that the 300W landmarks are different from the 68 landmarks obtained from FLAME.
The difference is primarily observed around the jawline.
\\
\textbf{Face Parsing:} Face parsing involves classifying the face into different regions (like nose, eyes, lips, etc.) at a pixel level.
We showcase our method's performance on two popular face parsing benchmarks - LaPa (\textbf{La}ndmark guided face \textbf{Pa}rsing) \cite{liu2020new} and CelebAMask-HQ \cite{CelebAMask-HQ}.
The faces in LaPa dataset are segmented into $11$ semantic classes.
CelebAMask-HQ is a subset of CelebA-HQ dataset containg $30,000$ high-resolution facial images wherein each image is segmented into various categories which include facial attributes like eyes, ears, nose, mouth, hair and skin.
In addition to facial attributes, this benchmarks also provides annotations for facial accessories like earrings, necklace, hats and sunglasses.\\
\textbf{Depth: } Using the depth maps curated by \cite{lee2021face}, we qualitatively showcase the results from our approach on depth estimation.

\subsection{Results}
We evaluate our approach on two facial analysis tasks - facial landmark localization and face parsing.
We also show qualitative results of our models in predicting facial depth. 
We compare our approach with existing methods that incorporate the use of synthetic data in 2 settings: a) using the entire sample set of $100k$ images for training, and b) using a randomly sampled batch of $10k$ images.
We show the performance of our approach in contrast to the popular PBR based method \cite{fitymi}, and, using DECA \cite{deca} to infer the annotations for the images from our dataset. 

Table \ref{tab:300w} showcases results for landmark localization task on the 300W dataset using the Normalized Mean Error (NME) metric - normalized by the inter-ocular outer eye distance.
We observe that our proposed synthetic backbone (SB) outperforms the PBR based method \cite{fitymi}.
The same trend is observed after label-finetuning.
Moreover, we also note that the performance of our approach is close to the state of the art methods trained entirely on real data for landmark localization.

Table \ref{tab:lapa} compares the results of our approach with other face parsing methods on the LAPA dataset.
We note that our label-finetuned synthetic backbone is very similar in performance to the PBR based method \cite{fitymi}.
We see a sharp increase in performance of our models when provided with the features $f_s$ from our synthetic feature backbone $\mathcal{E}$.
This is primarily seen in categories which are not a part of our annotation scheme like hair or have a considerable disparity from the segmentation maps in real world datasets - like eyebrows and lips.
Table \ref{tab:celeb} showcases results for face parsing on the CelebAMask-HQ dataset. We highlight that our approach is on-par with other state-of-the-art face parsing methods on this dataset. 

Fig. \ref{fig:lapa_res} shows the qualitative results on samples from the LaPa validation set for all the modalities. Notice that the predictions from the synthetic model are aligned to the groundtruth landmarks, despite never explicitly being trained on real samples. The predictions from the finetuned model also exhibit similar performance. The face parsing maps from out SB model are able to localize and segment different parts of the face accurately. The label finetuned model shows aligned predictions to the ground-truth annotations. Finally, we highlight that the predicted depth map by the SB model is qualitatively superior compared to the ground-truth reconstructed depth provided by the LaPa-D \cite{lee2021face} dataset. Notice how SB and SB+LF are able model the depth for glasses for the third identity. We showcase more qualitative results in the supplementary material.

\section{Conclusion and Limitations}
In conclusion, our work explores the integration of generative models for synthesizing high-quality face images and addresses the challenge of controllable annotations through the use of 3D Morphable Models (3DMMs). By leveraging generative synthetic data, coupled with dense multi-modal annotations, we demonstrate the efficacy of our approach in training a multi-task model for facial analysis. Our strategy includes label fine-tuning to enhance predictions on real-world datasets, supported by a robust dense multi-modal training pipeline with cross-task consistency and self-supervised losses. Our experiments reveal competitive performance against state-of-the-art single and multi-task models. Additionally, we contribute to the research community by releasing our curated dataset, pretrained models, and codebase, encouraging further exploration of synthetic data for real-world facial analysis tasks.\\
Limitations: The segmentation from the texture pipeline of the FLAME model introduces the need for a label finetuning step as FLAME does not have access to the hair, and other features like facial accessories. Moreover, the generative model used might not be consistent with the 3d annotations provided by FLAME in extreme poses.  We discuss in detail about this misalignment in the supplementary text.

\section*{Acknowledgement}
We gratefully acknowledge the support of Mercedes-Benz Research and Development India for this work, especially in the form of compute clusters to generate data and train models.
\small
\bibliographystyle{plain}
\bibliography{main}

\begin{thebibliography}{10}

\bibitem{eg3d}
Eric~R. Chan, Connor~Z. Lin, Matthew~A. Chan, Koki Nagano, Boxiao Pan, Shalini~De Mello, Orazio Gallo, Leonidas Guibas, Jonathan Tremblay, Sameh Khamis, Tero Karras, and Gordon Wetzstein.
\newblock Efficient geometry-aware {3D} generative adversarial networks.
\newblock In {\em arXiv}, 2021.

\bibitem{deng2019retinaface}
Jiankang Deng, Jia Guo, Zhou Yuxiang, Jinke Yu, Irene Kotsia, and Stefanos Zafeiriou.
\newblock Retinaface: Single-stage dense face localisation in the wild.
\newblock In {\em arxiv}, 2019.

\bibitem{ding2023diffusionrig}
Zheng Ding, Xuaner Zhang, Zhihao Xia, Lars Jebe, Zhuowen Tu, and Xiuming Zhang.
\newblock Diffusionrig: Learning personalized priors for facial appearance editing.
\newblock In {\em Proceedings of the IEEE/CVF Conference on Computer Vision and Pattern Recognition}, pages 12736--12746, 2023.

\bibitem{Falcon_PyTorch_Lightning_2019}
William Falcon and {The PyTorch Lightning team}.
\newblock {PyTorch Lightning}, March 2019.

\bibitem{deca}
Yao Feng, Haiwen Feng, Michael~J. Black, and Timo Bolkart.
\newblock Learning an animatable detailed {3D} face model from in-the-wild images.
\newblock {\em ACM Transactions on Graphics (ToG), Proc. SIGGRAPH}, 40(4):88:1--88:13, August 2021.

\bibitem{gomez2019large}
Francisco Gomez-Donoso, Sergio Orts-Escolano, and Miguel Cazorla.
\newblock Large-scale multiview 3d hand pose dataset.
\newblock {\em Image and Vision Computing}, 81:25--33, 2019.

\bibitem{ho2020denoising}
Jonathan Ho, Ajay Jain, and Pieter Abbeel.
\newblock Denoising diffusion probabilistic models, 2020.

\bibitem{hodavn2020bop}
Tom{\'a}{\v{s}} Hoda{\v{n}}, Martin Sundermeyer, Bertram Drost, Yann Labb{\'e}, Eric Brachmann, Frank Michel, Carsten Rother, and Ji{\v{r}}{\'\i} Matas.
\newblock Bop challenge 2020 on 6d object localization.
\newblock In {\em Computer Vision--ECCV 2020 Workshops: Glasgow, UK, August 23--28, 2020, Proceedings, Part II 16}, pages 577--594. Springer, 2020.

\bibitem{hodan2019photorealistic}
Tom{\'a}{\v{s}} Hoda{\v{n}}, Vibhav Vineet, Ran Gal, Emanuel Shalev, Jon Hanzelka, Treb Connell, Pedro Urbina, Sudipta Sinha, and Brian Guenter.
\newblock Photorealistic image synthesis for object instance detection.
\newblock {\em IEEE International Conference on Image Processing (ICIP)}, 2019.

\bibitem{HUANG2023272}
Wenjing Huang, Shikui Tu, and Lei Xu.
\newblock Ia-faces: A bidirectional method for semantic face editing.
\newblock {\em Neural Networks}, 158:272--292, 2023.

\bibitem{isola2018imagetoimage}
Phillip Isola, Jun-Yan Zhu, Tinghui Zhou, and Alexei~A. Efros.
\newblock Image-to-image translation with conditional adversarial networks, 2018.

\bibitem{jahanian2022generative}
Ali Jahanian, Xavier Puig, Yonglong Tian, and Phillip Isola.
\newblock Generative models as a data source for multiview representation learning, 2022.

\bibitem{Kar_2019_ICCV}
Amlan Kar, Aayush Prakash, Ming-Yu Liu, Eric Cameracci, Justin Yuan, Matt Rusiniak, David Acuna, Antonio Torralba, and Sanja Fidler.
\newblock Meta-sim: Learning to generate synthetic datasets.
\newblock In {\em Proceedings of the IEEE/CVF International Conference on Computer Vision (ICCV)}, October 2019.

\bibitem{karras2017progressive}
Tero Karras, Timo Aila, Samuli Laine, and Jaakko Lehtinen.
\newblock Progressive growing of gans for improved quality, stability, and variation.
\newblock {\em arXiv preprint arXiv:1710.10196}, 2017.

\bibitem{karras2019style}
Tero Karras, Samuli Laine, and Timo Aila.
\newblock A style-based generator architecture for generative adversarial networks.
\newblock In {\em Proceedings of the IEEE/CVF conference on computer vision and pattern recognition}, pages 4401--4410, 2019.

\bibitem{stylegan2}
Tero Karras, Samuli Laine, Miika Aittala, Janne Hellsten, Jaakko Lehtinen, and Timo Aila.
\newblock Analyzing and improving the image quality of stylegan.
\newblock In {\em Proceedings of the IEEE/CVF conference on computer vision and pattern recognition}, pages 8110--8119, 2020.

\bibitem{le2012interactive}
Vuong Le, Jonathan Brandt, Zhe Lin, Lubomir Bourdev, and Thomas~S Huang.
\newblock Interactive facial feature localization.
\newblock In {\em ECCV}, 2012.

\bibitem{CelebAMask-HQ}
Cheng-Han Lee, Ziwei Liu, Lingyun Wu, and Ping Luo.
\newblock Maskgan: Towards diverse and interactive facial image manipulation.
\newblock In {\em IEEE Conference on Computer Vision and Pattern Recognition (CVPR)}, 2020.

\bibitem{lee2021face}
Jihyun Lee, Binod Bhattarai, and Tae-Kyun Kim.
\newblock Face parsing from rgb and depth using cross-domain mutual learning.
\newblock In {\em Proceedings of the IEEE/CVF Conference on Computer Vision and Pattern Recognition}, pages 1501--1510, 2021.

\bibitem{li2018dsfd}
Jian Li, Yabiao Wang, Changan Wang, Ying Tai, Jianjun Qian, Jian Yang, Chengjie Wang, Jilin Li, and Feiyue Huang.
\newblock Dsfd: Dual shot face detector.
\newblock In {\em Proceedings of the IEEE Conference on Computer Vision and Pattern Recognition}, 2019.

\bibitem{flame}
Tianye Li, Timo Bolkart, Michael.~J. Black, Hao Li, and Javier Romero.
\newblock Learning a model of facial shape and expression from {4D} scans.
\newblock {\em ACM Transactions on Graphics, (Proc. SIGGRAPH Asia)}, 36(6):194:1--194:17, 2017.

\bibitem{mtpsl}
Wei-Hong Li, Xialei Liu, and Hakan Bilen.
\newblock Learning multiple dense prediction tasks from partially annotated data.
\newblock In {\em Proceedings of the IEEE/CVF Conference on Computer Vision and Pattern Recognition}, pages 18879--18889, 2022.

\bibitem{lin2021roi}
Yiming Lin, Jie Shen, Yujiang Wang, and Maja Pantic.
\newblock Roi tanh-polar transformer network for face parsing in the wild.
\newblock {\em Image and Vision Computing}, 112:104190, 2021.

\bibitem{liu2019end}
Shikun Liu, Edward Johns, and Andrew~J Davison.
\newblock End-to-end multi-task learning with attention.
\newblock In {\em Proceedings of the IEEE/CVF conference on computer vision and pattern recognition}, pages 1871--1880, 2019.

\bibitem{liu2020new}
Yinglu Liu, Hailin Shi, Hao Shen, Yue Si, Xiaobo Wang, and Tao Mei.
\newblock A new dataset and boundary-attention semantic segmentation for face parsing.
\newblock In {\em AAAI}, pages 11637--11644, 2020.

\bibitem{liu2015faceattributes}
Ziwei Liu, Ping Luo, Xiaogang Wang, and Xiaoou Tang.
\newblock Deep learning face attributes in the wild.
\newblock In {\em Proceedings of International Conference on Computer Vision (ICCV)}, December 2015.

\bibitem{DBLP:journals/corr/abs-1711-05101}
Ilya Loshchilov and Frank Hutter.
\newblock Fixing weight decay regularization in adam.
\newblock {\em CoRR}, abs/1711.05101, 2017.

\bibitem{nerf}
Ben Mildenhall, Pratul~P Srinivasan, Matthew Tancik, Jonathan~T Barron, Ravi Ramamoorthi, and Ren Ng.
\newblock Nerf: Representing scenes as neural radiance fields for view synthesis.
\newblock {\em Communications of the ACM}, 65(1):99--106, 2021.

\bibitem{newell2016stacked}
Alejandro Newell, Kaiyu Yang, and Jia Deng.
\newblock Stacked hourglass networks for human pose estimation.
\newblock In {\em Computer Vision--ECCV 2016: 14th European Conference, Amsterdam, The Netherlands, October 11-14, 2016, Proceedings, Part VIII 14}, pages 483--499. Springer, 2016.

\bibitem{ravi2020pytorch3d}
Nikhila Ravi, Jeremy Reizenstein, David Novotny, Taylor Gordon, Wan-Yen Lo, Justin Johnson, and Georgia Gkioxari.
\newblock Accelerating 3d deep learning with pytorch3d.
\newblock {\em arXiv:2007.08501}, 2020.

\bibitem{richter2016playing}
Stephan~R Richter, Vibhav Vineet, Stefan Roth, and Vladlen Koltun.
\newblock Playing for data: Ground truth from computer games.
\newblock In {\em Computer Vision--ECCV 2016: 14th European Conference, Amsterdam, The Netherlands, October 11-14, 2016, Proceedings, Part II 14}, pages 102--118. Springer, 2016.

\bibitem{roich2022pivotal}
Daniel Roich, Ron Mokady, Amit~H Bermano, and Daniel Cohen-Or.
\newblock Pivotal tuning for latent-based editing of real images.
\newblock {\em ACM Transactions on graphics (TOG)}, 42(1):1--13, 2022.

\bibitem{rombach2022highresolution}
Robin Rombach, Andreas Blattmann, Dominik Lorenz, Patrick Esser, and Björn Ommer.
\newblock High-resolution image synthesis with latent diffusion models, 2022.

\bibitem{unet}
Olaf Ronneberger, Philipp Fischer, and Thomas Brox.
\newblock U-net: Convolutional networks for biomedical image segmentation.
\newblock In {\em Medical Image Computing and Computer-Assisted Intervention--MICCAI 2015: 18th International Conference, Munich, Germany, October 5-9, 2015, Proceedings, Part III 18}, pages 234--241. Springer, 2015.

\bibitem{sagonas2013300}
Christos Sagonas, Georgios Tzimiropoulos, Stefanos Zafeiriou, and Maja Pantic.
\newblock 300 faces in-the-wild challenge: The first facial landmark localization challenge.
\newblock In {\em Proceedings of the IEEE international conference on computer vision workshops}, pages 397--403, 2013.

\bibitem{Sarkar_2023_CVPR}
Mausoom Sarkar, SR~Nikitha, Mayur Hemani, Rishabh Jain, and Balaji Krishnamurthy.
\newblock Parameter efficient local implicit image function network for face segmentation.
\newblock In {\em Proceedings of the IEEE/CVF Conference on Computer Vision and Pattern Recognition (CVPR)}, pages 20970--20980, June 2023.

\bibitem{next3d}
Jingxiang Sun, Xuan Wang, Lizhen Wang, Xiaoyu Li, Yong Zhang, Hongwen Zhang, and Yebin Liu.
\newblock Next3d: Generative neural texture rasterization for 3d-aware head avatars.
\newblock In {\em CVPR}, 2023.

\bibitem{te2021adaptive}
Gusi Te, Wei Hu, Yinglu Liu, Hailin Shi, and Tao Mei.
\newblock Adaptive graph representation learning and reasoning for face parsing.
\newblock {\em arXiv e-prints}, pages arXiv--2101, 2021.

\bibitem{te2020edge}
Gusi Te, Yinglu Liu, Wei Hu, Hailin Shi, and Tao Mei.
\newblock Edge-aware graph representation learning and reasoning for face parsing.
\newblock In {\em Computer Vision--ECCV 2020: 16th European Conference, Glasgow, UK, August 23--28, 2020, Proceedings, Part XII 16}, pages 258--274. Springer, 2020.

\bibitem{vandenhende2021multi}
Simon Vandenhende, Stamatios Georgoulis, Wouter Van~Gansbeke, Marc Proesmans, Dengxin Dai, and Luc Van~Gool.
\newblock Multi-task learning for dense prediction tasks: A survey.
\newblock {\em IEEE transactions on pattern analysis and machine intelligence}, 44(7):3614--3633, 2021.

\bibitem{varol17_surreal}
G{\"u}l Varol, Javier Romero, Xavier Martin, Naureen Mahmood, Michael~J. Black, Ivan Laptev, and Cordelia Schmid.
\newblock Learning from synthetic humans.
\newblock In {\em CVPR}, 2017.

\bibitem{wei2019accurate}
Zhen Wei, Si~Liu, Yao Sun, and Hefei Ling.
\newblock Accurate facial image parsing at real-time speed.
\newblock {\em IEEE Transactions on Image Processing}, 28(9):4659--4670, 2019.

\bibitem{fitymi}
Erroll Wood, Tadas Baltru{\v{s}}aitis, Charlie Hewitt, Sebastian Dziadzio, Thomas~J Cashman, and Jamie Shotton.
\newblock Fake it till you make it: face analysis in the wild using synthetic data alone.
\newblock In {\em Proceedings of the IEEE/CVF international conference on computer vision}, pages 3681--3691, 2021.

\bibitem{xu2023omniavatar}
Hongyi Xu, Guoxian Song, Zihang Jiang, Jianfeng Zhang, Yichun Shi, Jing Liu, Wanchun Ma, Jiashi Feng, and Linjie Luo.
\newblock Omniavatar: Geometry-guided controllable 3d head synthesis, 2023.

\bibitem{dadnet++}
Libing Zeng, Lele Chen, Wentao Bao, Zhong Li, Yi~Xu, Junsong Yuan, and Nima~Khademi Kalantari.
\newblock 3d-aware facial landmark detection via multi-view consistent training on synthetic data.
\newblock In {\em Proceedings of the IEEE/CVF Conference on Computer Vision and Pattern Recognition}, pages 12747--12758, 2023.

\bibitem{zhang2023adding}
Lvmin Zhang, Anyi Rao, and Maneesh Agrawala.
\newblock Adding conditional control to text-to-image diffusion models, 2023.

\bibitem{zheng2022decoupled}
Qingping Zheng, Jiankang Deng, Zheng Zhu, Ying Li, and Stefanos Zafeiriou.
\newblock Decoupled multi-task learning with cyclical self-regulation for face parsing.
\newblock In {\em Proceedings of the IEEE/CVF Conference on Computer Vision and Pattern Recognition}, pages 4156--4165, 2022.

\bibitem{zheng2021farl}
Yinglin Zheng, Hao Yang, Ting Zhang, Jianmin Bao, Dongdong Chen, Yangyu Huang, Lu~Yuan, Dong Chen, Ming Zeng, and Fang Wen.
\newblock General facial representation learning in a visual-linguistic manner.
\newblock {\em arXiv preprint arXiv:2112.03109}, 2021.

\bibitem{star}
Zhenglin Zhou, Huaxia Li, Hong Liu, Nanyang Wang, Gang Yu, and Rongrong Ji.
\newblock Star loss: Reducing semantic ambiguity in facial landmark detection.
\newblock In {\em Proceedings of the IEEE/CVF Conference on Computer Vision and Pattern Recognition (CVPR)}, pages 15475--15484, June 2023.

\end{thebibliography}

\end{document}